\documentclass{article}
\usepackage{spconf,amsmath,graphicx}
\usepackage{url}       
\usepackage{amssymb}   
\usepackage{amsthm}    
\usepackage{booktabs}  
\usepackage{siunitx}   
\usepackage{makecell}  
\usepackage{arydshln}  
\usepackage{textgreek} 

\usepackage{enumitem}
\setlist{nosep, leftmargin=14pt}

\usepackage{mwe} 

\title{Boundary-aware Instance Segmentation in Microscopy Imaging}
\name{
Thomas Mendelson$^{\star}$ \qquad
Joshua Francois$^{\dagger}$ \qquad
Galit Lahav$^{\dagger}$ \qquad
Tammy Riklin Raviv$^{\star}$
}
\address{$^{\star}$The School of Electrical and Computer Engineering, Ben-Gurion University of the Negev\\
$^{\dagger}$Department of Systems Biology, Harvard Medical School}
\begin{document}
%
\maketitle
\begin{abstract}
   Accurate delineation of individual cells in microscopy videos is essential for studying cellular dynamics, yet separating touching or overlapping instances remains a persistent challenge. Although foundation-model for segmentation such as SAM have broadened the accessibility of image segmentation, they still struggle to separate nearby cell instances in dense microscopy scenes without extensive prompting.

We propose a prompt-free, boundary-aware instance segmentation framework that predicts signed distance functions (SDFs) instead of binary masks, enabling smooth and geometry-consistent modeling of cell contours. A learned sigmoid mapping converts SDFs into probability maps, yielding sharp boundary localization and robust separation of adjacent instances. Training is guided by a unified Modified Hausdorff Distance (MHD) loss that integrates region- and boundary-based terms.

Evaluations on both public and private high-throughput microscopy datasets demonstrate improved boundary accuracy and instance-level performance compared to recent SAM-based and foundation-model approaches. Source code is available at\footnote{\url{https://github.com/ThomasMendelson/BAISeg.git}}.

\end{abstract}
\vspace{-.2cm}
\section{Introduction}
\label{sec:intro}
   Instance segmentation in live-cell microscopy is a critical task in biomedical image analysis, underpinning quantitative cell biology and clinical diagnostics. While semantic segmentation approaches can broadly separate foreground (cell regions) from background, they often fall short when cells are tightly clustered or overlapping~\cite{ronneberger2015unet, cai2020denseunet}. Instance segmentation methods aim to detect and delineate each individual cell, enabling downstream tasks such as tracking, morphological analysis, and phenotyping. However, cell segmentation in microscopy images remains challenging due to heterogeneous cell appearance, variability in shape and size, low contrast, and inhomogeneous illumination. These difficulties are further compounded in dense or dynamic scenes, where the primary bottleneck is the accurate separation of nearby or touching cell instances, whose contours often merge due to overlaps, diffuse boundaries, or complex interactions.
\begin{figure}[t]
  \centering
  \setlength{\tabcolsep}{1pt}
  \renewcommand{\arraystretch}{1.0}
  \begin{tabular}{cccc}    \includegraphics[width=.24\columnwidth,height=.24\columnwidth]{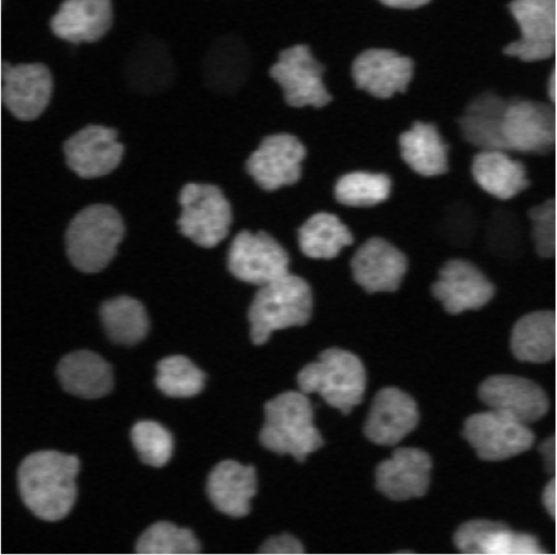} &   \includegraphics[width=.24\columnwidth,height=.24\columnwidth]{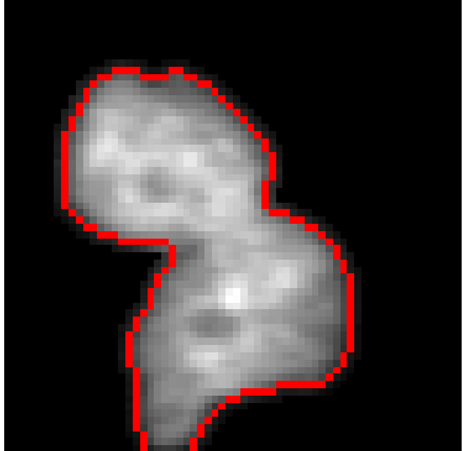} &
\includegraphics[width=.24\columnwidth,height=.24\columnwidth]{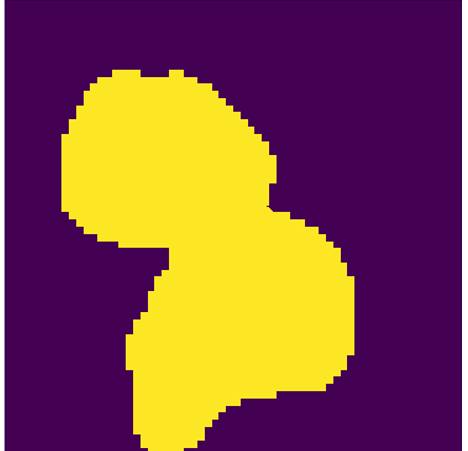} &
\includegraphics[width=.24\columnwidth,height=.24\columnwidth]{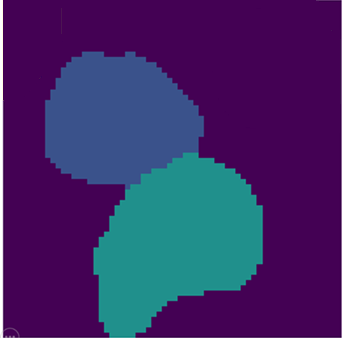} \\[-3pt]
    \scriptsize{Input Image} &
    \scriptsize{Zoom-in} &
    \scriptsize{GT Binary Map} &
    \scriptsize{GT Instance Labels}
  \end{tabular}
  \caption{\small
  \textbf{Examples illustrating the limitations of conventional pixel-wise loss functions for cell instance segmentation.}
  (a)~Original input image \(I\);
  (b)~Zoomed-in example from (a), showing an example edge prediction (in red) highlighting boundary ambiguity;
  (c)~Binary ground-truth segmentation map \(B_{\mathrm{GT}}\);
  (d)~Ground-truth instance label map \(\Gamma_{\mathrm{GT}}\).
  Pixel-wise losses fail to capture subtle boundary discrepancies in densely packed cellular regions.}
  \label{fig:ce_not_enough}
\end{figure}
Instance cell segmentation is an active field of study, see~\cite{maska2024csb} and references therein. 
The U-Net architecture~\cite{ronneberger2015unet} is a popular backbone thanks to its multi-scale structure and skip connections. Yet, while being excellent for segmentation in general, its training with the commonly used binary cross-entropy loss alone may fail in the presence of touching instances (see Fig.~\ref{fig:ce_not_enough}). The original U-Net framework proposed exactly for this problem incorporated a weighted loss to separate touching cells. Extensions introduced motion cues for time-lapse data~\cite{arbelle2018model,arbelle2022baseline}, explicit boundary-prediction heads, and post-processing such as watershed or distance-transform splitting~\cite{StarDist2018, Cellpose2021,liang2025cell}. Although effective in some settings, these approaches rely on hand-crafted heuristics or multi-stage pipelines, limiting robustness in dense cellular environments.

Recently, foundation models such as the Segment Anything Model (SAM)~\cite{kirillov2023segmentanything} have emerged as powerful general-purpose segmentation frameworks. Domain-specific adaptations like μSAM~\cite{luo2024segmentanythingmicroscopy} and Cellpose-SAM~\cite{stringer2025cellposesam} further tailor SAM to microscopy through fine-tuning or hybrid architectures. However, these approaches remain largely prompt-driven and may underperform in dense, instance-rich settings, where separating physically adjacent cells requires fine-grained boundary precision that interactive prompting cannot consistently provide.

To address these challenges and avoid prompting, we propose a boundary-aware instance segmentation framework that predicts a continuous Signed Distance Function (SDF) rather than a binary mask. A learned sigmoid maps the SDF to probabilistic segmentations, improving boundary precision and the separation of adjacent cells. 
Built upon U-Net’s multi-scale features~\cite{ronneberger2015unet}, our method incorporates instance separation directly into the loss via a differentiable Modified Hausdorff Distance (MHD) ~\cite{Dubuisson94}. The MHD loss quantifies discrepancies at the contour level rather than at the pixel level, eliminating the need for heuristic post-processing. Moreover, unlike boundary-classification methods~\cite{liang2025cell} that depend on watershed proposals and post-hoc merging, our approach enforces instance separation directly in the network.

We evaluated our method on both private and public high-throughput microscopy datasets, including challenging dense cellular scenes. Across all datasets, our approach was either the best or second-best performer when compared to Cellpose, SAM, and MicroSAM, as measured by a variant of the IoU score, widely used for instance-level accuracy.\\
\textbf{Contributions.} Our work (i) introduces a prompt-free SDF-based architecture for accurate contour localization and robust separation of touching instances, (ii) proposes a geometry-driven MHD loss that prioritizes contour fidelity over pixel-wise agreement, and (iii) demonstrates state-of-the-art performance—best or second-best among Cellpose, SAM, and MicroSAM—across private and public high-throughput microscopy datasets under the SEG metric.
\vspace{-.2cm}
\section{Method}
\label{sec:method}
    Our objective is to improve the separation of adjacent cells in dense microscopy images by predicting a continuous signed distance function (SDF) and optimizing a differentiable Modified Hausdorff Distance–based loss, enabling precise boundary localization and reduced topological errors.
\\
    \textbf{Boundary-based Loss.}  
    The proposed boundary-based loss is derived from the Hausdorff distance.  
    Let $S_{\mbox{\tiny{GT}}} \colon \Omega \to \{0,1\}$ and $S_{\mbox{\tiny{PRED}}} \colon \Omega \to \{0,1\}$ denote the binary GT segmentation of a single cell instance and the predicted segmentation, respectively, where $\Omega \subset \mathbb{R}^2$ is the entire image domain. Each segmentation mask can be represented as a set of pixels: $\omega_{\mbox{\tiny{GT}}} \triangleq \{\mathbf{x}_i\}_{i=1}^M$, with $\mathbf{x}_i \in \{\Omega \,|\, S_{\mbox{\tiny{GT}}}(\mathbf{x}_i) = 1\}$, and $\omega_{\mbox{\tiny{PRED}}} \triangleq \{\mathbf{y}_j\}_{j=1}^N$, with $\mathbf{y}_j \in \{\Omega \,|\, S_{\mbox{\tiny{PRED}}}(\mathbf{y}_j) = 1\}$. The cardinalities may differ, i.e., $M \triangleq |\omega_{\mbox{\tiny{GT}}}| \neq N \triangleq |\omega_{\mbox{\tiny{PRED}}}|$.
    
    The classical Hausdorff distance is defined as:
    \begin{equation}
    \label{eq:HDist}
    \begin{aligned}
    D_H(S_{\mbox{\tiny{GT}}}, S_{\mbox{\tiny{PRED}}}) = 
    \max \Big\{
    & \sup_{\mathbf{x}_i \in \omega_{\mbox{\tiny{GT}}}} \inf_{\mathbf{y}_j \in \omega_{\mbox{\tiny{PRED}}}} d(\mathbf{x}_i, \mathbf{y}_j),\\
    & \sup_{\mathbf{y}_j \in \omega_{\mbox{\tiny{PRED}}}} \inf_{\mathbf{x}_i \in \omega_{\mbox{\tiny{GT}}}} d(\mathbf{x}_i, \mathbf{y}_j)
    \Big\},
    \end{aligned}
    \end{equation}
    where $d(\mathbf{x}_i,\mathbf{y}_j)$ is the Euclidean distance between pixels $\mathbf{x}_i \in \omega_{\mbox{\tiny{GT}}}$ and $\mathbf{y}_j \in \omega_{\mbox{\tiny{PRED}}}$.
    Following~\cite{Dubuisson94}, we adopt an MHD by replacing the maximum and supremum operations in Eq.~\eqref{eq:HDist} with summations:
    \begin{equation}
    \label{eq:MHDist}
    \begin{aligned}
    D_{\mbox{\footnotesize{MH}}}(S_{\mbox{\tiny{GT}}}, S_{\mbox{\tiny{PRED}}}) =
    & \sum_{\mathbf{x}_i \in \omega_{\mbox{\tiny{GT}}}}
      \inf_{\mathbf{y}_j \in \omega_{\mbox{\tiny{PRED}}}}
      d(\mathbf{x}_i, \mathbf{y}_j) \\
    & + \sum_{\mathbf{y}_j \in \omega_{\mbox{\tiny{PRED}}}}
      \inf_{\mathbf{x}_i \in \omega_{\mbox{\tiny{GT}}}}
      d(\mathbf{x}_i, \mathbf{y}_j).
    \end{aligned}
    \end{equation}
    In the spirit of~\cite{riklin2014mhd}, we reformulate Eq.~\eqref{eq:MHDist} using SDFs. Let $\partial \omega \subset \Omega$ denote the boundary of $\omega$, representing an object region. The corresponding SDF $\phi \colon \Omega \to \mathbb{R}$ is defined as:
    \begin{equation}
    \label{eq:SDF}
    \phi(\mathbf{x}) =
    \begin{cases}
    d_E(\mathbf{x},\partial\omega), & \mathbf{x} \in \omega, \\
    -d_E(\mathbf{x},\partial\omega), & \mathbf{x} \in \Omega \setminus \omega,
    \end{cases}
    \end{equation}
    where $d_E(\cdot)$ is the Euclidean distance. We assume $\phi(\mathbf{x})$ is differentiable almost everywhere and satisfies the Eikonal equation $|\nabla \phi(\mathbf{x})| = 1$.
    
    Let $\phi_{\mbox{\tiny{GT}}}$ and $\phi_{\mbox{\tiny{PRED}}}$ be the SDFs for the GT and predicted segmentations, respectively, with boundaries $\partial\omega_{\mbox{\tiny{GT}}}$ and $\partial\omega_{\mbox{\tiny{PRED}}}$. The absolute value $|\phi_{\mbox{\tiny{GT}}}(\mathbf{x})|$ is the minimal Euclidean distance between $\mathbf{x}$ and the boundary of $S_{\mbox{\tiny{GT}}}$. Integrating $|\phi_{\mbox{\tiny{GT}}}(\mathbf{x})|$ over all $\mathbf{x} \in \partial\omega_{\mbox{\tiny{PRED}}}$ corresponds to the left-hand term of Eq.~\eqref{eq:MHDist}, and integrating $|\phi_{\mbox{\tiny{PRED}}}(\mathbf{x})|$ over $\mathbf{x} \in \partial\omega_{\mbox{\tiny{GT}}}$ corresponds to the right-hand term:
    
    \begin{equation}
    \label{eq:CMH}
    \begin{aligned}
    D_{\mbox{\footnotesize{CMH}}}(S_{\mbox{\tiny{GT}}}, S_{\mbox{\tiny{PRED}}}) =
    & \int_{\partial \omega_{\mbox{\tiny{GT}}}}
      \big| \phi_{S_{\mbox{\tiny{PRED}}}}(\mathbf{x}) \big| \, ds \\
    & + \int_{\partial \omega_{\mbox{\tiny{PRED}}}}
      \big| \phi_{S_{\mbox{\tiny{GT}}}}(\mathbf{x}) \big| \, ds.
    \end{aligned}
    \end{equation}
    where $ds$ denotes an infinitesimal curve (in 2D) element along the object boundary.
    
    In practice, we train a neural network to directly predict the SDF of the cell segmentation. To define a differentiable boundary-based loss in the spirit of Eq.~\eqref{eq:CMH}, we use a parameterized sigmoid function:
    \begin{equation}
    \sigma_{\alpha,\beta}(z) = \frac{1}{1 + \exp\left(-\left(\alpha z + \beta\right)\right)},
    \end{equation}
    where $\alpha$ and $\beta$ are learnable parameters. While $\phi_{\mbox{\tiny{PRED}}} \colon \Omega \to (-\infty,\infty)$, $\sigma_{\alpha,\beta}(\phi_{\mbox{\tiny{PRED}}})$ maps pixels to $[0,1]$, where each value can be interpreted as the probability of being inside the cell region.
    
    The predicted boundaries $\partial\omega_{\mbox{\tiny{PRED}}}$ are approximated by the differentiable soft boundary map:
    \begin{equation}
    \label{eq:BPred}
    B_{\mbox{\tiny{PRED}}}(\mathbf{x}) = \sigma_{\alpha,\beta}(\phi_{\mbox{\tiny{PRED}}}(\mathbf{x})) \, \sigma_{\alpha,\beta}(-\phi_{\mbox{\tiny{PRED}}}(\mathbf{x})),
    \end{equation}
    and similarly for the GT boundaries:
    \begin{equation}
    \label{eq:BGT}
    B_{\mbox{\tiny{GT}}}(\mathbf{x}) = \sigma_{\alpha,\beta}(\phi_{\mbox{\tiny{GT}}}(\mathbf{x})) \, \sigma_{\alpha,\beta}(-\phi_{\mbox{\tiny{GT}}}(\mathbf{x})).
    \end{equation}
    
\begin{figure*}[t]
    \centering
    \setlength{\tabcolsep}{2pt} 
    \renewcommand{\arraystretch}{1.0}
    \begin{tabular}{ccccc}
    \includegraphics[width=.19\textwidth]{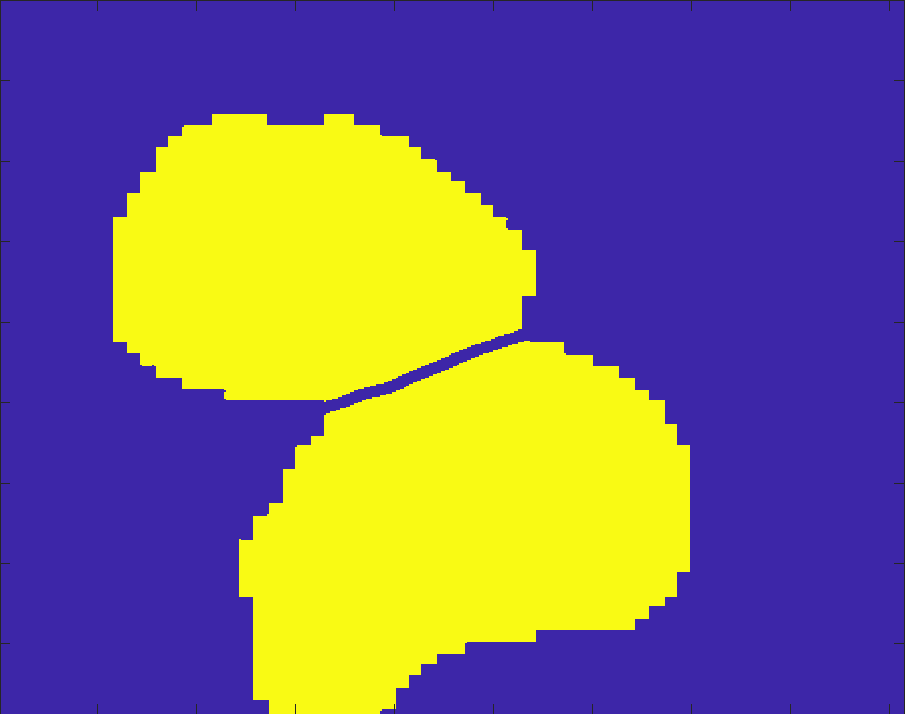} &
    \includegraphics[width=.19\textwidth]{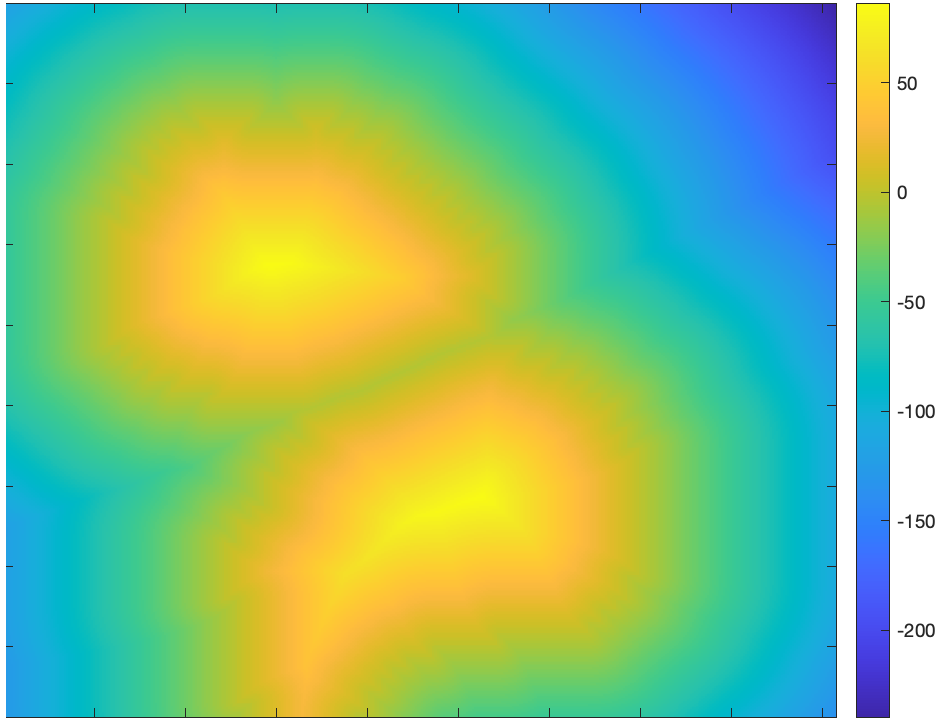} &
    \includegraphics[width=.19\textwidth]{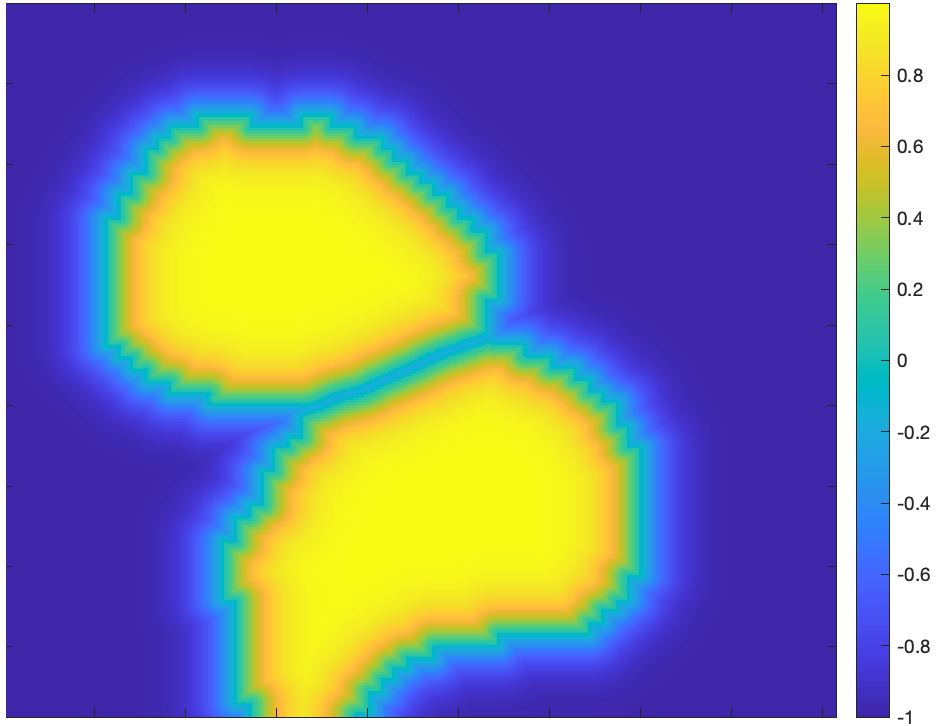} &
    \includegraphics[width=.19\textwidth]{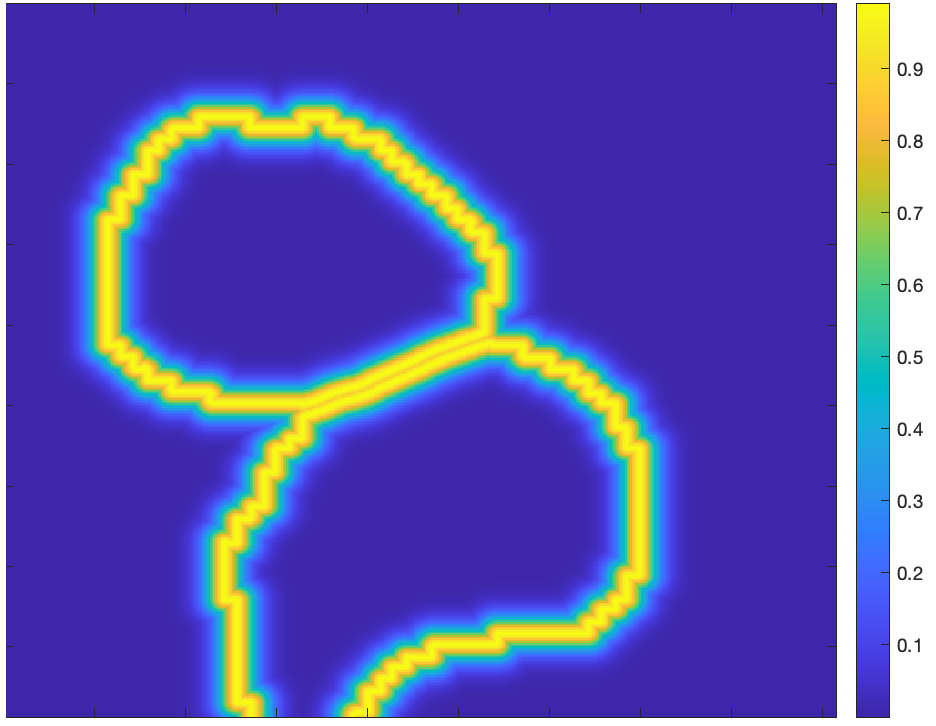} &
    \includegraphics[width=.19\textwidth]{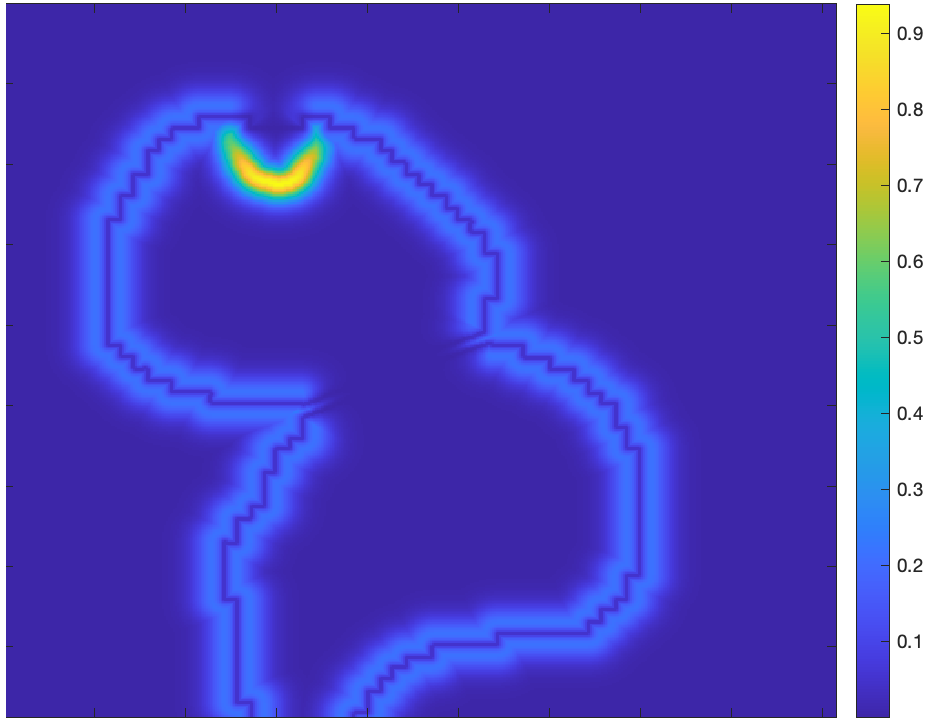} \\[-2pt]
    $S_{\mathrm{GT}}$ & $\phi_{\mathrm{GT}}$ &
    $\tanh_{\alpha,\beta}(\phi_{\mathrm{GT}})$ &
    $B_{\mathrm{GT}}$ &
    $|\tanh_{\mathrm{GT}}|\!\odot\!B_{\mathrm{PRED}}$ \\[4pt]
    \includegraphics[width=.19\textwidth]{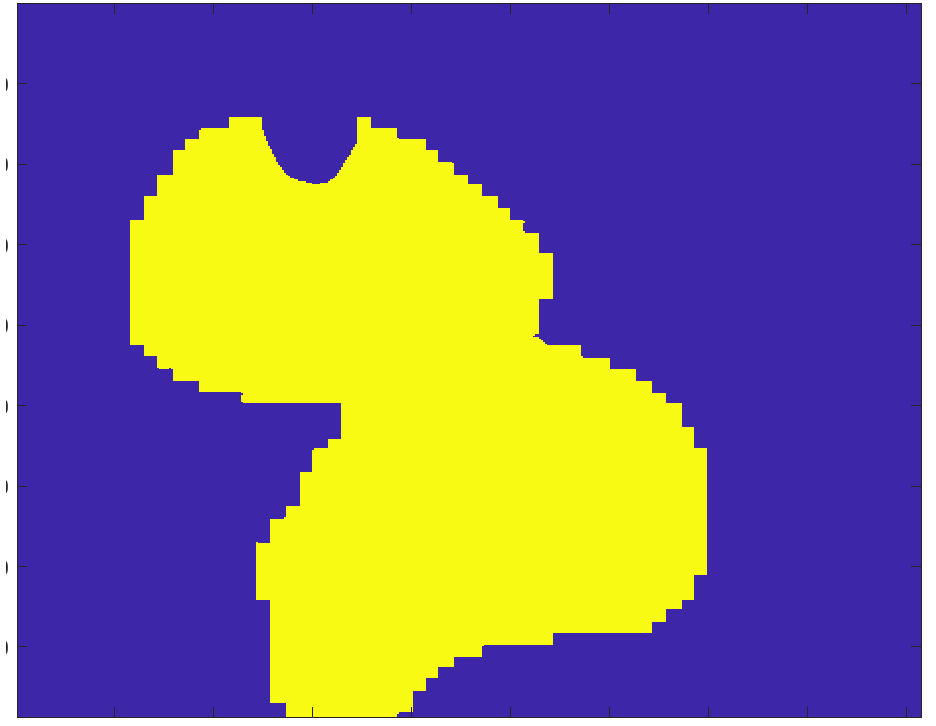} &
    \includegraphics[width=.19\textwidth]{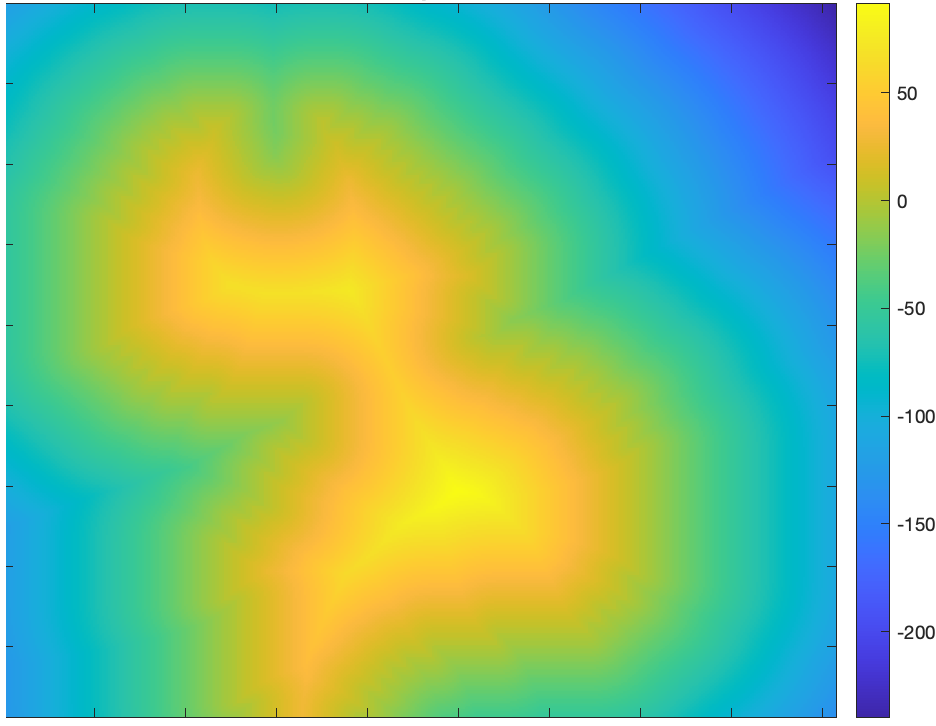} &
    \includegraphics[width=.19\textwidth]{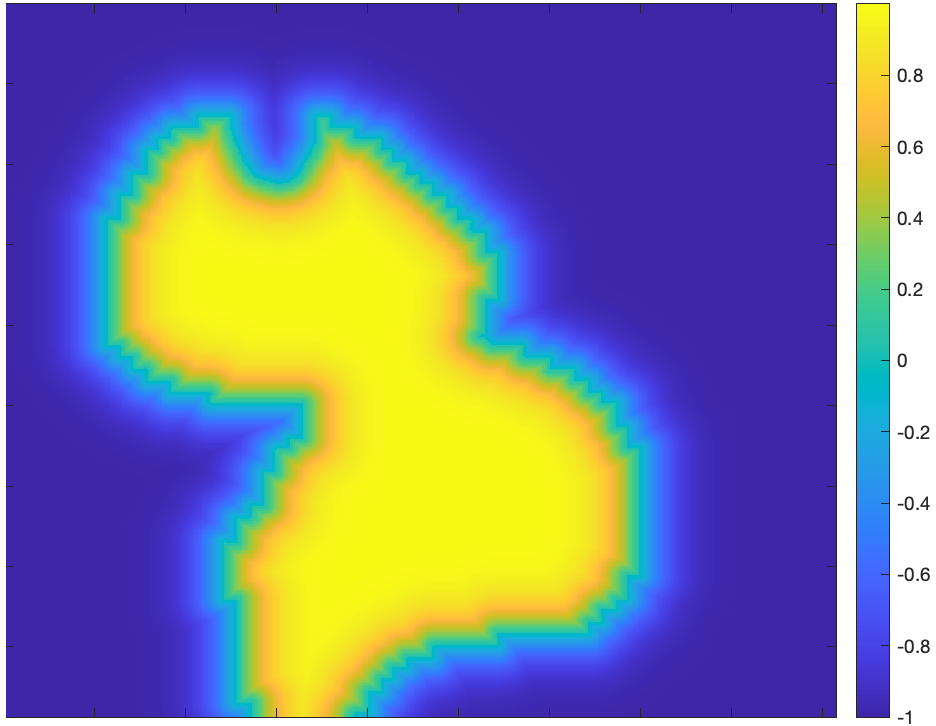} &
    \includegraphics[width=.19\textwidth]{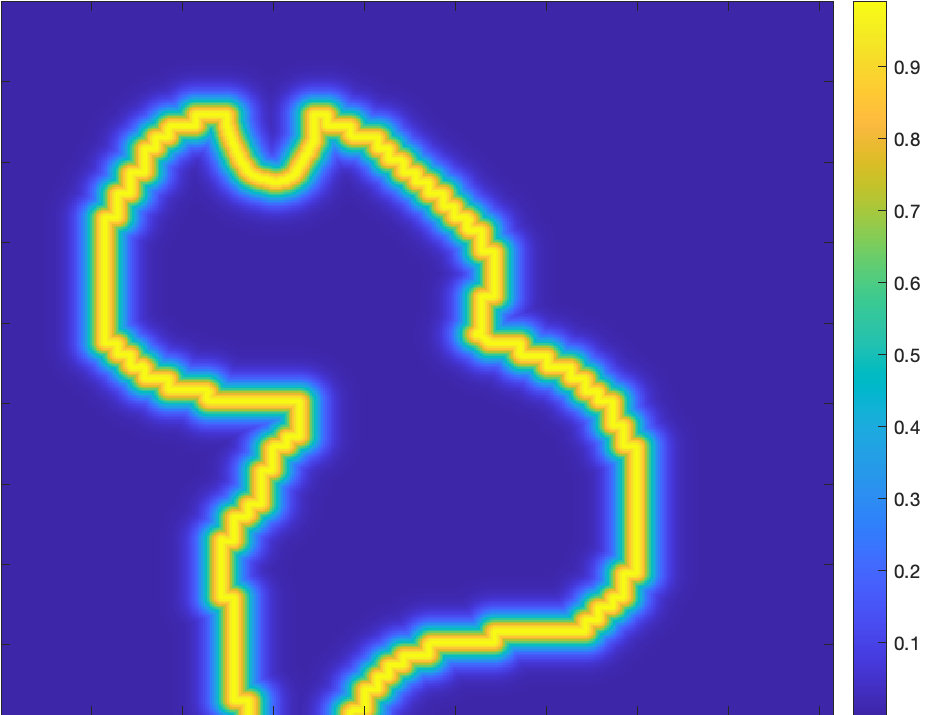} &
    \includegraphics[width=.19\textwidth]{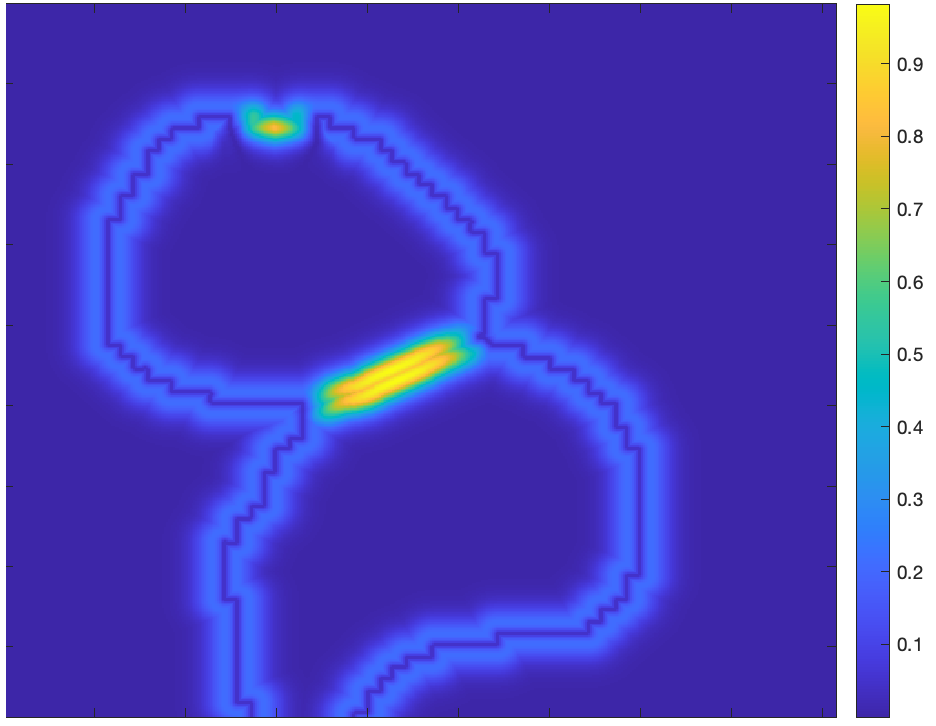} \\[-2pt]
    $S_{\mathrm{PRED}}$ & $\phi_{\mathrm{PRED}}$ &
    $\tanh_{\alpha,\beta}(\phi_{\mathrm{PRED}})$ &
    $B_{\mathrm{PRED}}$ &
    $|\tanh_{\mathrm{PRED}}|\!\odot\!B_{\mathrm{GT}}$
    \end{tabular}
    \caption{\small \textbf{Visualization of the boundary-based loss computation.}
    The binary GT segmentation $S_{\mathrm{GT}}$ is derived from the GT label maps, where touching cells are separated in a preprocessing stage using known cell boundaries.
    The network predicts the signed distance function (SDF) of the cell segmentation map, $\phi_{\mathrm{PRED}}$; the corresponding binary mask $S_{\mathrm{PRED}}$ is shown for reference.
    For comparison, the GT SDF, $\phi_{\mathrm{GT}}$, is also displayed.
    Soft boundary maps, $B_{\mathrm{GT}}$ and $B_{\mathrm{PRED}}$, are obtained as $\sigma(\phi)\,\sigma(-\phi)$.
    The rightmost images illustrate Eq.~\ref{eq:leftSDF} (top) and Eq.~\ref{eq:rightSDF} (bottom).
    Boundary-based loss values are computed by summing over the respective image domains.}
    \label{fig:BoundaryBased}
    \end{figure*}

    Given the soft boundary maps' formulation (Eq.~\ref{eq:BGT}, Eq.~\ref{eq:BPred}), we define two MHD loss terms that discretize and approximate the left- and right-hand sides of Eq.~\eqref{eq:CMH}. Since SDF values can be arbitrarily scaled in $\mathbb{R}$, we apply a hyperbolic tangent, which monotonically maps $(-\infty,\infty)$ to $[-1,1]$:

    \begin{equation}
    \label{eq:leftSDF}
    \begin{aligned}
    \mathcal{L}_{\mbox{\tiny{LMHD}}}
    &= \sum_{\mathbf{x} \in \Omega}
       B_{\mbox{\tiny{GT}}}(\mathbf{x}) \,
       \big| \tanh_{\alpha,\beta}(\phi_{\mbox{\tiny{PRED}}}(\mathbf{x})) \big| \\
    &= B_{\mbox{\tiny{GT}}} \cdot
       \big| \tanh_{\mbox{\tiny{PRED}}} \big|.
    \end{aligned}
    \end{equation}
    
    \begin{equation}
    \label{eq:rightSDF}
    \begin{aligned}
    \mathcal{L}_{\mbox{\tiny{RMHD}}}
    &= \sum_{\mathbf{x} \in \Omega}
       B_{\mbox{\tiny{PRED}}}(\mathbf{x}) \,
       \big| \tanh_{\alpha,\beta}(\phi_{\mbox{\tiny{GT}}}(\mathbf{x})) \big| \\
    &= B_{\mbox{\tiny{PRED}}} \cdot
       \big| \tanh_{\mbox{\tiny{GT}}} \big|.
    \end{aligned}
    \end{equation}
    The parameterized hyperbolic tangent can be formulated by:
    \begin{equation}
    \label{eq:tanh}
    \tanh_{\alpha,\beta}(z) = \tanh\big( \alpha (z - \beta) \big) = 2\,\sigma_{\alpha,\beta}(z) - 1.
    \end{equation}
    \textbf{Unified loss function.}
Rather than predicting binary masks, our framework outputs continuous SDFs by using two additional loss terms. First, we introduce a Least Squares Error (LSE) loss between the predicted and ground-truth SDFs after applying the parameterized hyperbolic tangent:
    \begin{equation}
    \label{eq:LSE}
    \mathcal{L}_{\mbox{\tiny{LSE}}} =
    \left[ \tanh_{\alpha,\beta}(\phi_{\mbox{\tiny{PRED}}}) -
           \tanh_{\alpha,\beta}(\phi_{\mbox{\tiny{GT}}}) \right]^2.
    \end{equation}
    Second, we define an SDF-based cross-entropy (CE) loss using the probabilistic interpretation of the predicted segmentation maps, i.e., $\sigma_{\alpha,\beta}(\phi_{\mbox{\tiny{PRED}}})$:
    \begin{equation}
    \label{eq:CE}
    \begin{aligned}
    \mathcal{L}_{\mbox{\tiny{CE}}}
    &= - \Big[
    S_{\mbox{\tiny{GT}}} \,
       \log \sigma_{\alpha,\beta}(\phi_{\mbox{\tiny{PRED}}})
    \\[-2pt]
    &\quad + \big(1 - S_{\mbox{\tiny{GT}}}\big)
       \log \big(1 - \sigma_{\alpha,\beta}(\phi_{\mbox{\tiny{PRED}}})\big)
    \Big].
    \end{aligned}
    \end{equation}
    \noindent
    The final unified objective combines all four terms:
    \begin{equation}
    \label{eq:UnifiedLoss}
    \mathcal{L}_{\mbox{\tiny{Total}}} =
    \lambda_{\mbox{\tiny{LMHD}}} \mathcal{L}_{\mbox{\tiny{LMHD}}}
    + \lambda_{\mbox{\tiny{RMHD}}} \mathcal{L}_{\mbox{\tiny{RMHD}}}
    + \lambda_{\mbox{\tiny{LSE}}} \mathcal{L}_{\mbox{\tiny{LSE}}}
    + \lambda_{\mbox{\tiny{CE}}} \mathcal{L}_{\mbox{\tiny{CE}}},
    \end{equation}
    where $\lambda_{\mbox{\tiny{LMHD}}}, \lambda_{\mbox{\tiny{RMHD}}}, \lambda_{\mbox{\tiny{LSE}}},$ and $\lambda_{\mbox{\tiny{CE}}}$ are weighting coefficients.
    An illustration of the intermediate outputs is shown in Fig.~\ref{fig:BoundaryBased}.
    \\
\textbf{Implementation Details.}
For all experiments (except SAM) we used a U-Net backbone with feature widths \{128, 256, 512, 1024\} (approximately 49.7M learnable parameters) trained on 2D patches of size \(128\times128\) with a batch size of 16. Standard augmentations—random affine transforms~\cite{ronneberger2015unet}, random crops, flips, and moderate brightness/contrast variations—are applied during training. The learnable sigmoid parameters are initialized as \(\alpha{=}4\) and \(\beta{=}0\). Optimization uses AdamW with a learning rate of \(1\!\times\!10^{-4}\) and weight decay \(1\!\times\!10^{-3}\). The unified loss (Eq.~\ref{eq:UnifiedLoss}) is weighted by \(\lambda_{\mathrm{LMHD}}{=}0.9\), \(\lambda_{\mathrm{RMHD}}{=}10^{-1}\), \(\lambda_{\mathrm{LSE}}{=}1\), and \(\lambda_{\mathrm{CE}}{=}1\).
Training was performed on a single NVIDIA V100 GPU taking approximately 45 minutes per dataset including augmentation.
Inference required on average 0.77 seconds per image.

\section{Experiments and Results}
\label{sec:experiments}
\vspace{-.2cm}
\subsection{Datasets}
\label{ssec:datasets}
Both public and private datasets were used for evaluation. \\
\textbf{CSB Dataset.} For public data, we use four 2D sequences from the Cell Segmentation Benchmark (CSB)~\cite{maska2014benchmark,maska2024csb,ulman2017ctc}\footnote{\url{https://celltrackingchallenge.net/}}: Fluo-N2DH-SIM+, Fluo-N2DH-HeLa, Fluo-N2DH-GOWT1, and Fluo-C2DL-MSC, spanning challenging real fluorescence and brightfield imaging as well as high-throughput simulated data. Segmentations for the CSB datasets are provided as gold (expert-annotated, though partial) and silver masks, the latter obtained via label fusion of outputs from multiple top-performing algorithms.\\
\textbf{Spheroid Dataset.}
The private dataset was provided by the Lahav Lab and consists of 3D irradiated MCF7 human breast cancer spheroids acquired with Selective Plane Illumination Microscopy (SPIM) across five post-irradiation timepoints. This volumetric dataset contains densely packed nuclei, dynamic morphological changes, and heterogeneous p21 expression. Ground-truth segmentations were generated manually using \textit{napari}\footnote{\url{https://napari.org}}~\cite{napari2023}.\\
\textbf{Preprocessing.} Across all datasets, we remove 1-pixel-wide label borders introduced by annotation tools, while preserving all true cell regions and overlaps, to obtain clean instance separability for SDF computation.\\
\textbf{Training-test split.}
The CSB benchmark provides two annotated training sequences per dataset; We therefore trained on the first sequence and evaluated on the second. For the 3D irradiated spheroid dataset, we used the manual 3D annotations to extract 539 labeled 2D slices, splitting them into 70\% for training and 30\% for testing, ensuring that slices from the same z-stack were not split across sets.
\vspace{-.2cm}
\subsection{Evaluation Measure}
The SEG measure, introduced by the CSB organizers, quantifies segmentation accuracy as the mean Jaccard index between each ground-truth object and its best-matching predicted object
where a prediction is considered a valid match to a ground-truth cell segmentation only if its intersection is higher than 0.5. 
\vspace{-.2cm}
\subsection{Ablation Study: MHD Loss Components}
We evaluated the contribution of each loss term on the Fluo-N2DH-GOWT1 and Fluo-C2DL-MSC datasets by selectively enabling or disabling individual components of the total loss. As shown in Table~\ref{tab:mhd_loss}, using both MHD terms (\(\mathcal{L}_{\mathrm{LMHD}}\) and \(\mathcal{L}_{\mathrm{RMHD}}\)) consistently yields the highest SEG scores, confirming their importance for enforcing accurate boundary alignment and improving instance separation.

\begin{table}[t]
    \centering
    \footnotesize
    \caption{\textbf{SEG scores for different loss combinations.}} 
    \resizebox{\linewidth}{!}{%
    \begin{tabular}{cccc|cc}
        \toprule
        $\mathcal{L}_{\mathrm{CE}}$ &
        $\mathcal{L}_{\mathrm{LSE}}$ &
        $\mathcal{L}_{\mathrm{LMHD}}$ &
        $\mathcal{L}_{\mathrm{RMHD}}$ &
        \textbf{GOWT1} &
        \textbf{MSC} \\
        \midrule
        \checkmark & \checkmark &            &            & \underline{.901(.180)} & \underline{.759(.123)} \\
        \checkmark & \checkmark &            & \checkmark & .882(.045)              & .634(.147)              \\
        \checkmark & \checkmark & \checkmark &            & .899(.020)              & .689(.123)              \\
        \checkmark & \checkmark & \checkmark & \checkmark & \textbf{.944(.018)}     & \textbf{.775(.102)}     \\
        \bottomrule
    \end{tabular}}
    \label{tab:mhd_loss}
\end{table}
\vspace{-.2cm}
\subsection{Experiments}

We evaluate our method on all five datasets described in Section~\ref{ssec:datasets}, using the same set of baselines—UNet, SAM (ViT\_B), $\mu$SAM, and Cellpose-SAM—for both public and private data. Quantitative segmentation accuracy (SEG) is summarized in Table~\ref{tab:combined_results}, and qualitative visualizations for the CSB datasets and the 3D spheroid slices are shown in Fig.~\ref{fig:seg-results}. Across all datasets, our boundary-aware SDF-based model achieves the best or second-best performance, despite being fully prompt-free.
\begin{table}[t]
    \centering
    \footnotesize
    \setlength{\tabcolsep}{3pt}
    \caption{\textbf{SEG (\(\uparrow\)) performance across all datasets.}
    Mean(std) over test images. A dashed line separates the 3D spheroid dataset from the four CSB datasets.
    For SAM (ViT\_B), $p{=}32$ and $p{=}64$ indicate prompt-grid densities of $32{\times}32$ and $64{\times}64$. 
    $\mu$SAM results ($^{\ast}$) are computed on 20 test images due to high inference cost.
    }

    \begin{tabular}{lc:cccc}
    \toprule
    \textbf{Method} &
    \textbf{MCF7} &
    \textbf{SIM+} &
    \textbf{HeLa} &
    \textbf{GOWT1} &
    \textbf{MSC} \\
    \midrule
    UNet & .492(.120) & .629(.254) & .774(.026) & .904(.025) & .420(.194) \\
    SAM ($p{=}32$) & .411(.177) & .508(.256) & .795(.042) & \underline{.943(.019)} & .229(.140) \\
    SAM ($p{=}64$) & .431(.182) & .518(.243) & \textbf{.917(.012)} & .938(.015) & .253(.141) \\
    \textit{$\mu$SAM} & .517(.184) & \textbf{.789(.104)}\rlap{$^{\ast}$} & .867(.022)\rlap{$^{\ast}$} & .843(.030)\rlap{$^{\ast}$} & \underline{.713(.107)}\rlap{$^{\ast}$} \\
    Cellpose & \underline{.521(.106)} & .597(.206) & .849(.044) & .933(.026) & .554(.175) \\
    Ours & \textbf{.798(.093)} & \underline{.788(.077)} & \underline{.892(.030)} & \textbf{.944(.018)} & \textbf{.775(.102)} \\
    \bottomrule
    \end{tabular}
    \label{tab:combined_results}
\end{table}

\begin{figure}[t]
    \centering
    \setlength{\tabcolsep}{1.5pt}
    \begin{tabular}{cccc}
        \includegraphics[width=2.1cm,height=2.1cm]{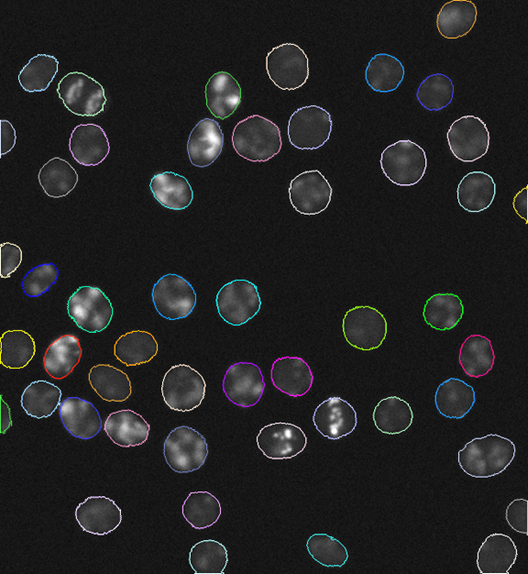} &
        \includegraphics[width=2.1cm,height=2.1cm]{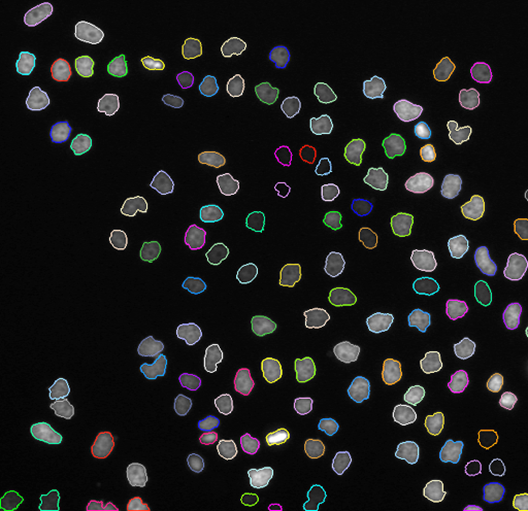} &
        \includegraphics[width=2.1cm,height=2.1cm]{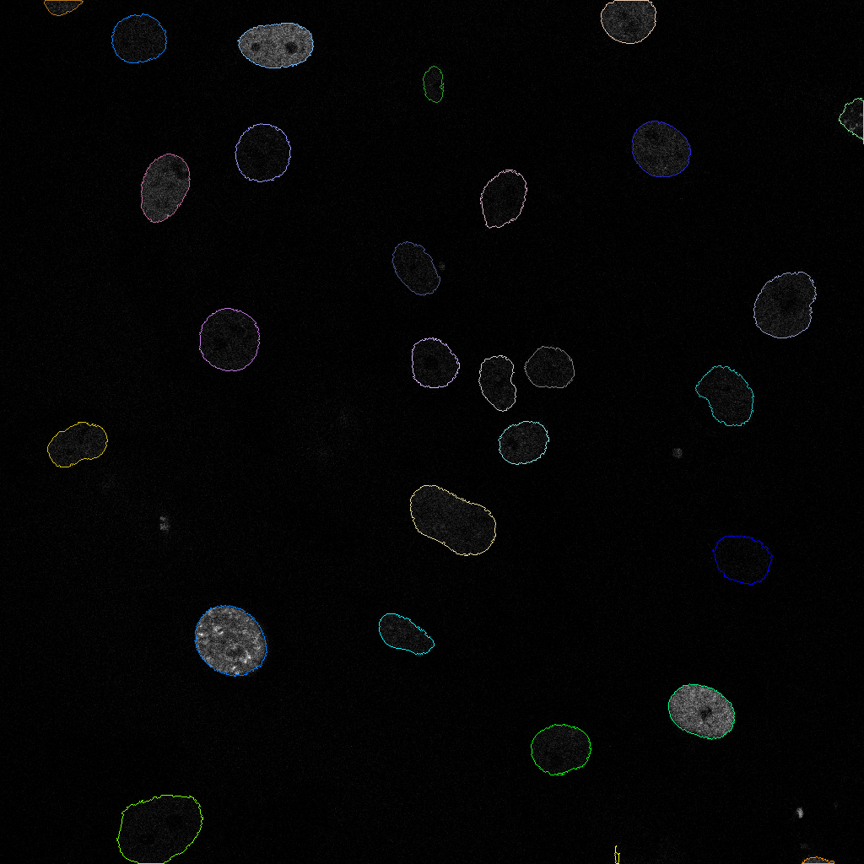} &
        \includegraphics[width=2.1cm,height=2.1cm]{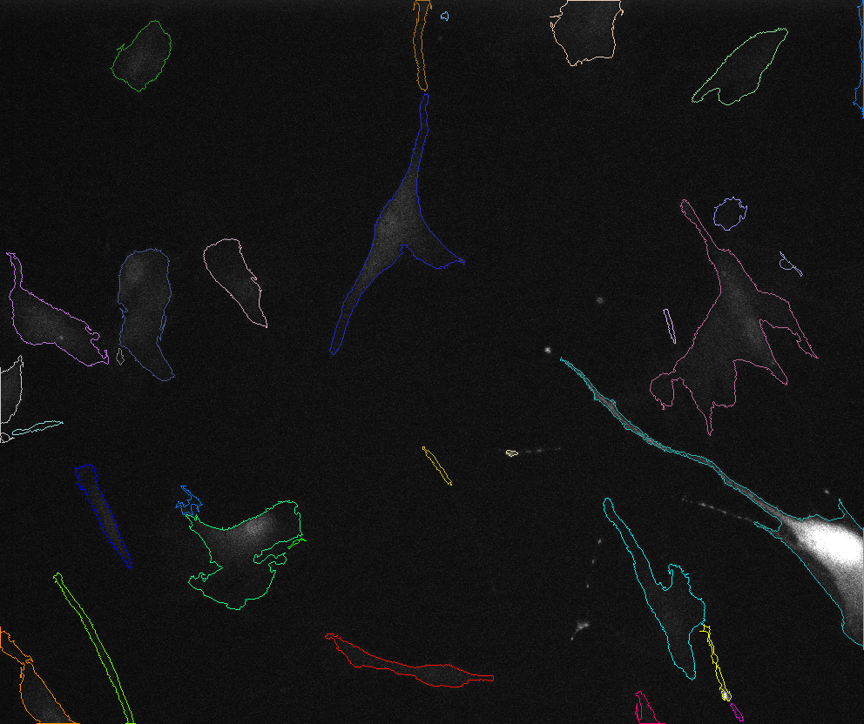} \\
        \footnotesize{N2DH-GOWT1} & \footnotesize{C2DL-MSC}& \footnotesize{N2DH-SIM+} & \footnotesize{N2DL-HeLa}\\
         \includegraphics[width=2.1cm,height=2.1cm]{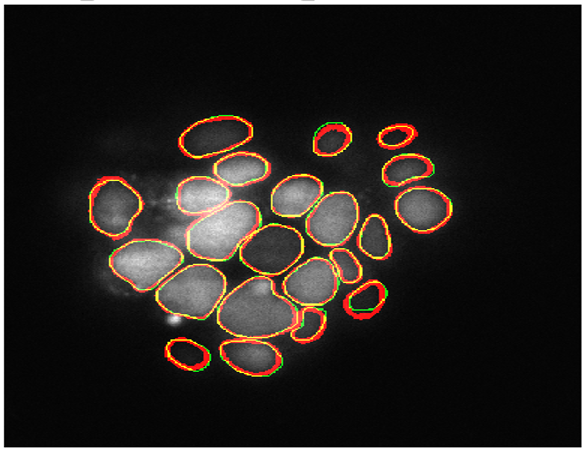} &
        \includegraphics[width=2.1cm,height=2.1cm]{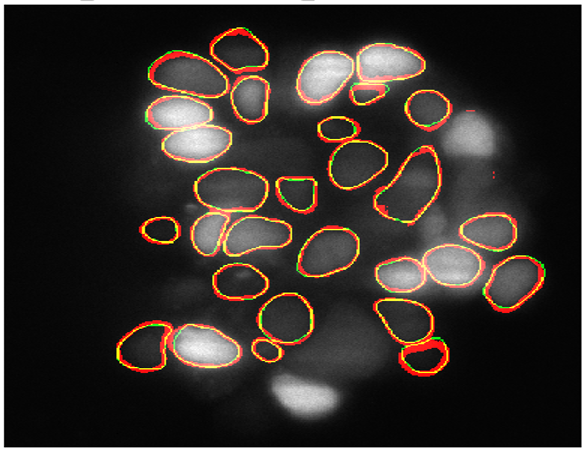} &
        \includegraphics[width=2.1cm,height=2.1cm]{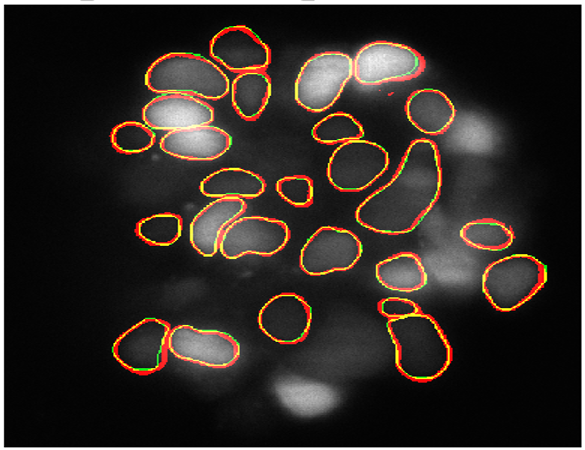} &
        \includegraphics[width=2.1cm,height=2.1cm]{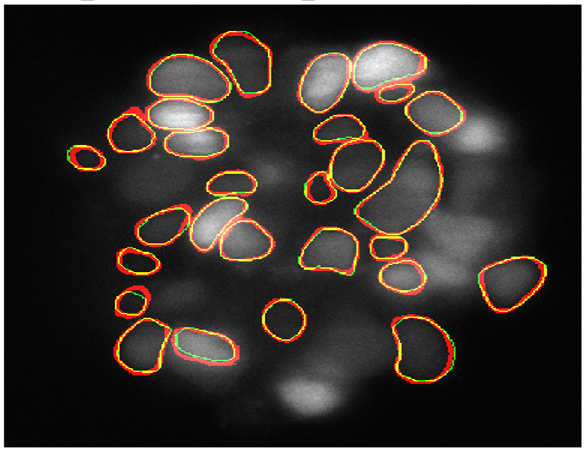}\\
         \includegraphics[width=2.1cm,height=2.1cm]{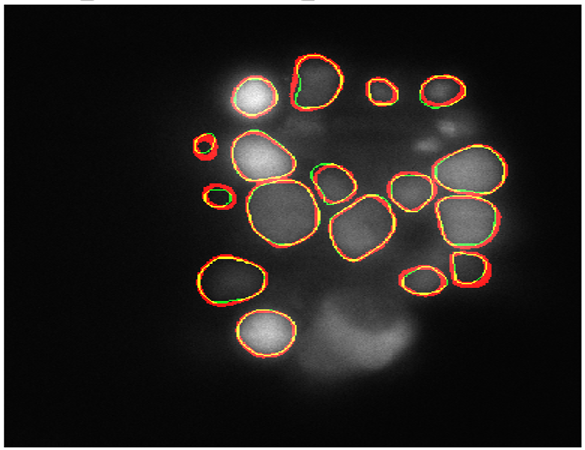} &
        \includegraphics[width=2.1cm,height=2.1cm]{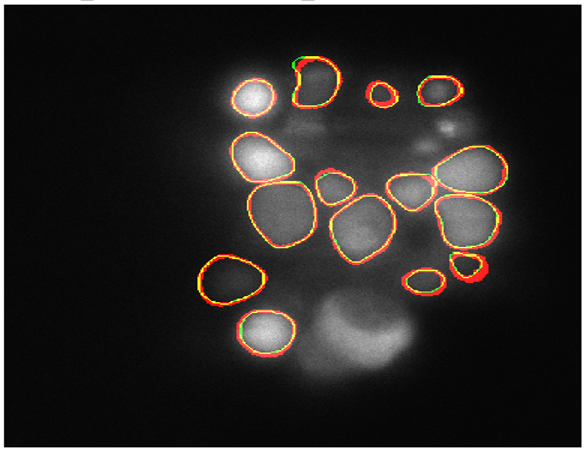} &
        \includegraphics[width=2.1cm,height=2.1cm]{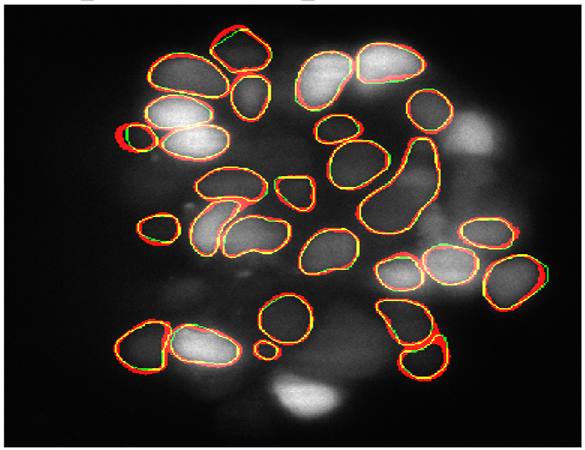} &
        \includegraphics[width=2.1cm,height=2.1cm]{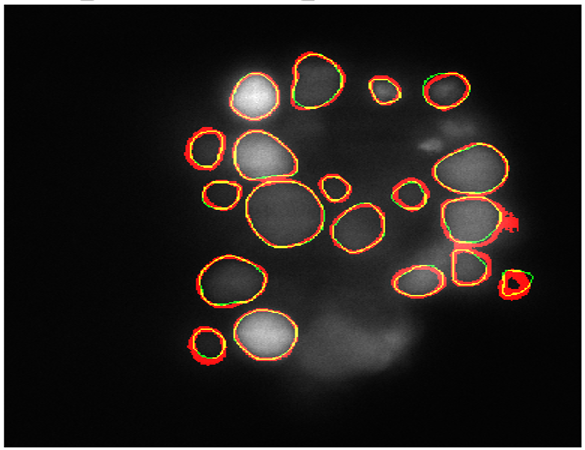}\\
        \multicolumn{4}{c}{MCF7}
    \end{tabular}
    \caption{\textbf{Qualitative segmentation results} Top four images - CSB datasets, bottom two rows - representative 2D slices from the private spheroid data.}
    \label{fig:seg-results}
\end{figure}
\vspace{-.2cm}
\section{Summary}
\label{sec:summary}
We introduced a boundary-aware instance segmentation framework that predicts continuous SDFs and optimizes a differentiable MHD loss to achieve precise boundary localization and reliable separation of closely packed cells. Experiments on both the CSB datasets and a 3D irradiated tumor spheroid dataset demonstrate consistent improvements over existing methods, yielding superior boundary accuracy and instance-level segmentation. 
Future work will extend this approach to 3D+t data and incorporate temporal consistency for improved tracking.
\clearpage
\pagebreak
\section{Compliance with Ethical Standards}
This study used publicly available microscopy datasets and additional spheroid microscopy data provided by co-author Joshua Francois (Harvard Medical School).  
The Harvard data were collected under institutional ethical approval and shared for analysis; no new experiments involving human participants or animals were performed by the authors.

\section{Acknowledgments}
\label{sec:acknowledgments}
This study was supported by the United States–Israel Binational Science Foundation (BSF 2019/135) and was partially supported by the National Institutes of Health (NIH) grant R35 GM139572.


\bibliographystyle{IEEEbib}
\bibliography{references}

\end{document}